\RequirePackage{booktabs}
\documentclass[pdflatex,sn-mathphys-num]{sn-jnl}

\usepackage{graphicx}%
\usepackage{multirow}%
\usepackage{amsmath,amssymb,amsfonts}%
\usepackage{amsthm}%
\usepackage{mathrsfs}%
\usepackage[title]{appendix}%
\usepackage{xcolor}%
\usepackage{textcomp}%
\usepackage{manyfoot}%
\usepackage{booktabs}%
\usepackage{algorithm}%
\usepackage{algorithmicx}%
\usepackage{algpseudocode}%
\usepackage{listings}%
\usepackage{subcaption}
\usepackage{rotating}
\usepackage{multicol}
\usepackage{titlesec}

\theoremstyle{thmstyleone}%
\theoremstyle{thmstyletwo}%

\theoremstyle{thmstylethree}%

\titleformat{\paragraph}
{\normalfont\normalsize\bfseries}{\theparagraph}{1em}{}
\titlespacing*{\paragraph}
{0pt}{3.25ex plus 1ex minus .2ex}{1.5ex plus .2ex}

\begin{document}


\title[Explainability Methods for Hardware Trojan Detection]{Explainability Methods for Hardware Trojan Detection: A Systematic Comparison}


\author*[1]{\fnm{Paul} \sur{Whitten}}\email{pcw@case.edu}

\author[1]{\fnm{Francis} \sur{Wolff}}\email{fxw12@case.edu}
\equalcont{These authors contributed equally to this work.}

\author[1]{\fnm{Chris} \sur{Papachristou}}\email{cap2@case.edu}
\equalcont{These authors contributed equally to this work.}

\affil*[1]{\orgdiv{Electrical, Computer, and Systems Engineering}, \orgname{Case Western Reserve University}, \orgaddress{\street{10900 Euclid Avenue}, \city{Cleveland}, \postcode{44106}, \state{Ohio}, \country{USA}}}


\abstract{

Hardware trojans are malicious circuits which compromise the functionality and
security of an integrated circuit (IC). These circuits are manufactured directly
into the silicon and cannot be fixed by security patches like software. The
solution would require a costly product recall by replacing the IC and hence,
early detection in the design process is essential. Hardware detection
at best provides statistically based solutions with many false positives and false
negatives. These detection methods require more thorough explainable analysis
to filter out false indicators.


Existing explainability methods developed for general domains like image classification do not always provide the actionable insights hardware engineers need. A question remains: how do domain-aware property analysis, model-agnostic case-based reasoning, and model-agnostic feature attribution techniques compare for hardware security applications?

This work compares three categories of explainability for gate-level hardware
trojan detection on the Trust-Hub benchmark dataset: (1) domain-aware
property-based analysis of 31 circuit-specific features derived from gate
fanin patterns, flip-flop distances, and primary Input/Output (I/O) connectivity; (2)
model-agnostic case-based reasoning using k-nearest neighbors for precedent-based
explanations; and (3) model-agnostic feature attribution methods (Local Interpretable 
Model-agnostic Explanations (LIME), SHapley Additive exPlanations (SHAP), gradient)
that provide generic importance scores without circuit-level context.

The findings show that different explainability approaches offer distinct advantages
for hardware security practitioners. The domain-aware
property-based method analyzes 31 circuit properties. Detection decisions are explained through familiar concepts like ``high fanin complexity near outputs
indicates potential trojan triggers.'' Case-based reasoning (k-nearest neighbors) achieves 96.51\% correspondence between predictions and training exemplars. In contrast, LIME and SHAP show only moderate per-gate agreement (Spearman $\rho = 0.30$ mean over $n = 11{,}392$ gates, 95\% bootstrap CI $[0.29, 0.31]$; global concatenated $\rho = 0.31$, $p \ll 10^{-300}$), and yield generic feature importance scores that lack circuit-level context for validation or remediation planning.
Detection performance using eXtreme Gradient Boosting (XGBoost) classification with optimized threshold
achieves 48.08\% precision and 69.44\% recall on 11,392 held-out test gates (F1~=~0.568, Matthews Correlation Coefficient (MCC)~=~0.575,
area under the precision-recall curve (AUPRC)~=~0.637) at the optimized threshold of 0.940. This represents a 4.25-fold precision improvement over
a support vector machine (SVM) baseline reimplemented under identical experimental conditions (11.33\% precision, 70.83\% recall,
F1~=~0.195 at threshold 0.050) with a 7.4-fold reduction in false-positive density (4.74 vs.\ 35.0 FP per 1,000 gates).
Random Forest achieves comparable F1 (0.555, 58.46\% precision, 52.78\% recall) with half the
false-positive density of XGBoost (2.37 vs.\ 4.74 FP per 1,000 gates), confirming that the explainable AI (XAI) findings
generalize across classifier choices.

Additionally, gradient-based feature
attribution (Simonyan et al., 2013), with a speedup factor of 7 over SHAP, yields
the same model-agnostic feature weights as SHAP and LIME, confirming that
computational efficiency alone cannot substitute for domain-aware
interpretability.

This work provides empirical evidence comparing domain-aware property analysis,
model-agnostic case-based reasoning, and model-agnostic feature attribution techniques
for hardware security applications. The results show that property-based and case-based
approaches offer complementary advantages: domain alignment and precedent-based
interpretability, respectively, compared to generic feature rankings. These findings
have implications for XAI deployment across domains where practitioners must
validate and act on Machine Learning (ML) predictions.

}

\keywords{explainable artificial intelligence, hardware trojan detection, domain-aware explainability, model-agnostic explanations, circuit property analysis, LIME, SHAP, XGBoost, Trust-Hub benchmark, Case-explainer}



\maketitle

\section{Introduction}\label{sec1}

When a machine learning system flags a gate in a billion-transistor IC as a potential hardware trojan, the security engineer needs an explanation in circuit terms. A bare ``feature importance: 0.73'' does little to support validation or remediation. Without interpretations grounded in circuit-design principles, practitioners cannot separate real alerts from false positives, verify suspect structural patterns against domain knowledge, or translate detections into security actions.

Hardware trojans are malicious circuit modifications introduced during design or fabrication. Unlike software bugs, they are permanent in silicon, activate through rare input combinations engineered to evade conventional test (often 1 in $2^{64}$ states), and operate at the gate level with no software-visible trace, potentially compromising confidentiality, integrity, or availability.

Machine learning models flag suspicious circuit structures from gate-level netlists, but opaque-box predictions lack the interpretable justifications hardware security engineers need. General-purpose XAI techniques such as LIME~\cite{ribeiro2016should} and SHAP~\cite{lundberg2017unified} explain predictions through feature perturbation and game-theoretic attribution; they are domain-agnostic and broadly applicable, but produce generic importance scores. Domain-aware methods, in contrast, encode circuit-specific knowledge (fanin patterns, flip-flop connectivity, proximity to primary I/O) directly into the explanation, aligning with how hardware engineers already reason about circuits but requiring domain effort to design. \textit{Do model-agnostic explainability techniques developed for general ML applications provide the actionable insights hardware security practitioners need, or do domain-aware methods tailored to circuit characteristics offer better interpretability?} Prior work has not answered this empirically.

This work provides such an empirical comparison on the Trust-Hub benchmark, contrasting domain-specific circuit-property analysis with general-purpose feature attribution.

The primary contributions are:

\begin{enumerate}
    \item \textbf{Domain-Aware Property Analysis}: A property-based
    explainability method that analyzes 31 gate-level circuit-specific features
    grounded in hardware design principles. It produces explanations such as,
    ``High LGFi = 12 at 2 levels upstream combined with low FFo = 1 matches
    rare-event trigger patterns,'' which engineers can validate using familiar
    circuit-analysis concepts.

    \item \textbf{Case-Based Reasoning with k-Nearest Neighbors}: An inherently
    interpretable predictor ($k = 5$) achieves 96.51\% correspondence between
    predictions and training exemplars, justifying detections by precedent:
    ``this gate's feature profile matches 4 of 5 similar training cases that
    were trojans.''

    \item \textbf{Systematic Quantitative Comparison with Model-Agnostic Baselines}: M2--M5 use a single XGBoost classifier with five different post-hoc explanation backends, isolating the effect of the explanation mechanism from the underlying detector. LIME and SHAP show moderate per-gate agreement (Spearman $\rho = 0.30$ over $n = 11{,}392$ gates, 95\% bootstrap CI $[0.29, 0.31]$; global concatenated $\rho = 0.31$, $p \ll 10^{-300}$): significantly correlated but far from identical. Gradient-based attribution~\cite{simonyan2013deep} is also evaluated for computational efficiency.

    \item \textbf{Improved Detection Performance with XGBoost and Random Forest}: XGBoost achieves 48.08\% precision at 69.44\% recall (F1~=~0.568, MCC~=~0.575, AUPRC~=~0.637); Random Forest achieves 58.46\% precision at 52.78\% recall (F1~=~0.555, MCC~=~0.553, AUPRC~=~0.512). Both substantially improve over the Hasegawa SVM baseline reimplemented under identical conditions (11.33\% precision, 70.83\% recall, F1~=~0.195): 4.25$\times$ precision (XGBoost) and 7.4$\times$ lower false-positive density (4.74 vs.\ 35.0 FP per 1{,}000 gates).
\end{enumerate}

Section~\ref{background} covers hardware trojans and XAI background. Section~\ref{sec:related} reviews related work. Sections~\ref{methods} and~\ref{sec:experimental} present methodology and experimental setup. Section~\ref{results} reports results. Section~\ref{sec:limitations} discusses limitations and future work. Section~\ref{conclusion} concludes.

\section{Background}\label{background}

\subsection {Hardware Trojans}

\begin{figure}[H]
    \centering
    \includegraphics[width=9.0cm, alt={A taxonomy of hardware trojans}]{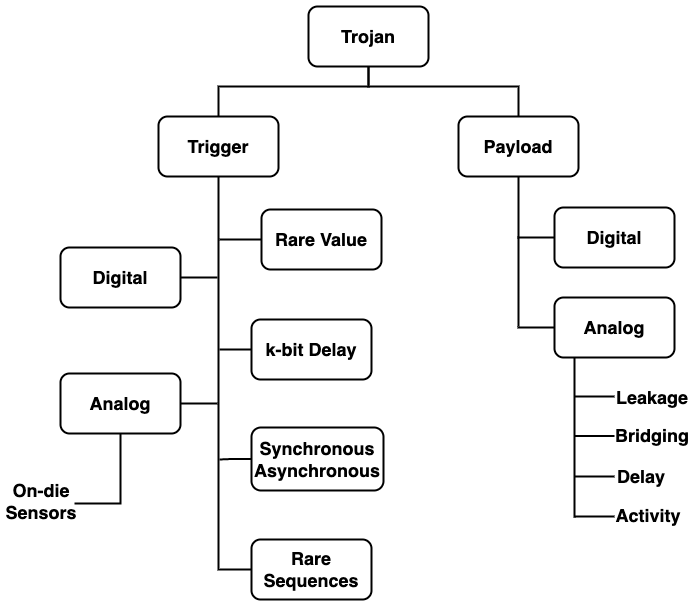}
    \caption{Trojan circuit taxonomy\cite{4484928}.}
    \label{fig:hw_trojan_circuit_taxonomy}
\end{figure}

Cost-driven outsourcing of integrated-circuit fabrication creates an opening for hardware trojans that weaken confidentiality, integrity, or availability of trusted ICs. Figure~\ref{fig:hw_trojan_circuit_taxonomy} shows the standard taxonomy: every trojan consists of a trigger and a payload, each of which may be digital or analog, with rare values, $k$-bit delays, or rare input sequences as typical activation conditions and control/status/data alteration or analog leakage as typical exploits~\cite{4484928}.

\begin{figure}[H]
    \centering
    \includegraphics[width=9.0cm, alt={A confidentiality rare event hardware trojan}]{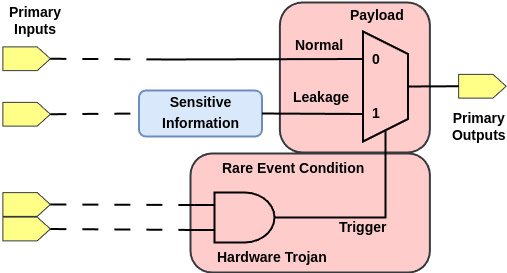}
    \caption{Confidentiality rare event hardware trojan model\cite{whitten2024naecon}.}
    \label{fig:cre_impact_model}
\end{figure}

Figure~\ref{fig:cre_impact_model} shows a confidentiality trojan implemented in gate-level logic: a rare-event AND trigger conditioned on primary inputs gates a multiplexer payload that leaks sensitive data to a primary output. Detectability rests on the trigger being rare (e.g., one of $2^n$ input combinations) and the payload reaching an observable output, so static gate-level analysis can flag suspect subcircuits by fanin complexity and reachability to primary outputs~\cite{4484928, 7604700, hasegawa2020hardware, px6s-sm21-22}.

Hasegawa et al.\ \cite{7604700, hasegawa2020hardware} encode this intuition as five per-net features used throughout this work: Logic Gate Fanins (LGFi, inputs to logic gates two levels upstream), Flip-Flop Input (FFi, upstream logic levels to a flip-flop), Flip-Flop Output (FFo, downstream logic levels to a flip-flop), Primary Input (PI, upstream levels to the nearest primary input), and Primary Output (PO, downstream levels to the nearest primary output). Their original SVM addresses class imbalance via dynamic weighting; subsequent work uses synthetic oversampling~\cite{10444240} and other rebalancing strategies but does not improve the interpretability of the resulting decisions.

\subsection{Trust-Hub Benchmark Dataset}

The Trust-Hub benchmark~\cite{6657085, shakya2017benchmarking, slayback2015computer, px6s-sm21-22} provides gate-level netlists with labeled trojan insertions. This work uses 30 digital combinational circuits drawn from RS232 communication peripherals and ISCAS benchmarks; analog and transistor-level circuits are out of scope.

\section{Related Work}\label{sec:related}

\subsection{Machine Learning for Hardware Trojan Detection}

\subsubsection{Hasegawa et al. Baseline}

Hasegawa et al. established a foundational approach for ML-based hardware trojan
detection from gate-level netlists \cite{7604700, hasegawa2020hardware}. Their
method extracts five structural features from each net: LGFi, FFi, FFo, PI,
and PO. Using Support Vector
Machine (SVM) classification with dynamic class weighting to address data
imbalance, they reported 82.6\% recall on the Trust-Hub benchmark dataset. 

Precision computed from their published confusion matrices is
5.13\%. Approximately 95\% of gates flagged as trojans are false positives. Their published confusion matrices (Table~V, dynamic weighting)
show an average true negative rate of 49.5\%, corresponding to a false positive rate
of approximately 50.5\%. The sheer volume of false positives limits practical
utility for real-world deployment where security engineers must manually inspect
falsely flagged gates.

To provide an apples-to-apples comparison, this work reimplements the Hasegawa
SVM baseline under identical experimental conditions: the same Trust-Hub dataset,
same five structural features, and same 60/20/20 random split. The reimplemented
SVM (radial basis function (RBF) kernel, \texttt{class\_weight=`balanced'}) achieves 11.33\% precision and
70.83\% recall at the val-optimal threshold of 0.050 (F1~=~0.195, MCC~=~0.274,
35.0 FP per 1,000 gates). XGBoost achieves 48.08\% precision at threshold 0.940,
a 4.25-fold precision improvement with a 7.4-fold reduction in false-positive density
(4.74 vs.\ 35.0 FP per 1,000 gates). Random Forest further reduces the FP burden to
2.37 FP per 1,000 gates (14.8-fold below the SVM baseline) at 58.46\% precision.

\subsubsection{Recent Advances}

Rathor and Rastogi's HT-Pred system \cite{rathor2025ht} uses 479 Trust-Hub
circuits with 605 structural and functional features, substantially larger than
prior published datasets. Their Deep Neural Network achieves 98.95\%
accuracy for circuit-level binary classification (trojan-infected vs.
trojan-free circuits). However, their work addresses a fundamentally different
problem than this work: they classify entire circuits rather than localizing
specific trojan gates within netlists. Circuit-level classification benefits
from balanced datasets (approximately 1:1 ratio of infected to clean circuits)
and provides limited actionability, as security engineers still must manually
search through hundreds or thousands of gates to locate the actual trojan
components.

In contrast, this work performs gate-level (node-level) trojan localization on
heavily imbalanced datasets (approximately 158 to one ratio of benign to trojan
gates), directly identifying suspicious gates to enable targeted remediation.
While HT-Pred achieves higher overall accuracy on the circuit-level
task, the gate-level localization provides the specific information needed for
practical trojan removal or mitigation.

\subsubsection{Ensemble Methods for Hardware Trojan Detection}

Negishi and Togawa~\cite{10444240} compare Random Forest, AdaBoost, and Gradient
Boosting against SVM on Trust-Hub, finding ensemble methods consistently improve
F-measure on the imbalanced detection task. Method~1 in this work uses an
ensemble of 31 XGBoost classifiers, each trained on a distinct subset of the
five Hasegawa features rather than on resampled data, aggregated by
$E_{PARS}$-weighted voting. The subset structure yields property-level
interpretability that homogeneous ensembles (Random Forest, AdaBoost) do not
provide.

\subsubsection{Class Imbalance Handling in Hardware Trojan Detection}

Trust-Hub gate-level netlists carry a 158:1 benign-to-trojan ratio overall and
157:1 in the held-out test fold. Classifiers trained without imbalance handling
bias toward the majority class. Standard responses include reweighting
\cite{king2001logistic}, synthetic oversampling, and cost-sensitive learning.
This work uses XGBoost's \texttt{scale\_pos\_weight} set to
$N_{\text{benign}}/N_{\text{trojan}}$, following King and
Zeng's~\cite{king2001logistic} weighting prescription, and achieves 69.44\%
recall on the 157:1 test fold.

Table~\ref{tab:related_work_comparison} positions the present work against
representative prior gate-level trojan detection methods on the Trust-Hub
benchmark. Same-protocol rows use identical splits and thresholding; published
rows are included for context and are not directly comparable.

\begin{sidewaystable}
\centering
\caption{Gate-level trojan detection on the Trust-Hub benchmark across
representative methods. ``Same-protocol'' rows use this work's 60/20/20
random split, threshold selected on the validation set, and the 11{,}392-gate
held-out test set. ``Published'' rows quote headline figures from the cited
papers under each paper's own protocol. These figures are not directly
comparable to the same-protocol rows.}
\label{tab:related_work_comparison}
\small
{\begin{tabular*}{\textheight}{@{\extracolsep\fill}p{3.8cm}p{2.5cm}p{2.5cm}p{1.1cm}p{1.1cm}p{0.9cm}p{4.1cm}}
\hline
\textbf{Method} & \textbf{Classifier} & \textbf{Protocol} & \textbf{P (\%)} & \textbf{R (\%)} & \textbf{F1} & \textbf{Source} \\
\hline
Hasegawa et al.~\cite{7604700} & SVM (RBF) & Published (LOCO) & 5.13 & 82.6 & 0.097 & \cite{7604700} Table~V \\
Hasegawa et al.\ (reimpl.) & SVM (RBF) & Same-protocol & 11.33 & 70.83 & 0.195 & This work \\
This work & XGBoost & Same-protocol & 48.08 & 69.44 & 0.568 & This work \\
This work & RF & Same-protocol & 58.46 & 52.78 & 0.555 & This work \\
SALTY~\cite{mahfuz2025salty} & Graph attention & Published (LOFO) & ---$^{d}$ & 98.47$^{d}$ & ---$^{d}$ & \cite{mahfuz2025salty} (125-dim graph features) \\
GNN4HT~\cite{yasaei2025gnn} & GNN & Published & --- & --- & --- & No P/R/F1 reported~\cite{yasaei2025gnn} \\
\hline
\multicolumn{7}{p{0.95\textheight}}{$^{d}$SALTY reports TPR/TNR (98.47\%/98.14\%) on 125-dim local graph features under leave-one-family-out cross-validation; precision and F1 not reported. Published rows use different features and protocols and are not directly comparable to same-protocol rows. LOCO: leave-one-circuit-out cross-validation; LOFO: leave-one-family-out cross-validation.}
\end{tabular*}}
\end{sidewaystable}

The same-protocol rows establish the comparison of interest: under identical
splits, thresholding, and metrics, the gradient-boosted detector substantially
improves precision over the Hasegawa SVM baseline (48.08\% vs.\ 11.33\%, a
4.25-fold improvement) at comparable recall. Graph-based detectors operate on
richer local features and are not directly comparable on this five-feature
representation. Their published numbers are included to situate the line of
work rather than to claim parity.

\subsection{Explainable AI Techniques}
\subsubsection{General XAI Methods}

Feature-importance methods identify which input features drive predictions,
globally or locally; example-based methods such as case-based reasoning
\cite{Caruana1999CasebasedEO} retrieve similar training instances as
explanations; attribution methods (LIME~\cite{ribeiro2016should},
SHAP~\cite{lundberg2017unified}) assign per-feature importance scores, with
SHAP offering local-accuracy and consistency guarantees at higher compute cost.

\subsubsection{XAI for Hardware Security}

Explainable AI applied to gate-level trojan detection remains thin. A systematic
search over 2016--2025 publications (queries: ``explainable AI hardware trojan,''
``interpretable machine learning hardware security,'' ``XAI hardware trojan
detection,'' ``explainability gate-level netlist analysis'') surfaces no prior
work that systematically compares domain-aware against model-agnostic XAI
methods on the Trust-Hub benchmark. Most ML-based trojan detection uses
opaque-box models without explainability
\cite{7604700, hasegawa2020hardware, rathor2025ht, 10444240}; when explanations
appear they are typically Random-Forest feature-importance rankings without
cross-method evaluation.

The most directly related XAI works:
Pan and Mishra~\cite{pan2022hardware} apply SHAP and LIME to hardware-malware
detection over performance-counter features, a different feature space, but
the same XAI-auditing motivation.
Mahfuz et al.\ (SALTY)~\cite{mahfuz2025salty} couple a graph neural network with
post-hoc XAI refinement and report
TPR~=~98.47\%, TNR~=~98.14\% on a 125-dimensional local feature set under
leave-one-family-out cross-validation, illustrating the headroom available from
richer representations.
Yasaei et al.~\cite{yasaei2025gnn} show that GNN feature representations improve
cross-circuit generalization over the five-feature Hasegawa baseline.
Ma et al.~\cite{ma2025hardware} extend GNN-based detection with harmonic
centrality on Trust-Hub. Su et al.~\cite{su2025lut} propose an explainable GNN
for LUT-level trojan localization in FPGA netlists. None of these systematically
compare XAI methods on a fixed gate-level structural feature representation.

This work fills that gap by comparing five explainability methods spanning
domain-aware property analysis, model-agnostic case-based reasoning, and
model-agnostic feature attribution (LIME, SHAP, gradient) on classification
performance, explanation consistency, and computational efficiency.

\subsection{Explainable Artificial Intelligence}

\begin{figure}[H]
    \centering
    \includegraphics[width=11.0cm, alt={Traditional ML interactions constrasted to XAI}]{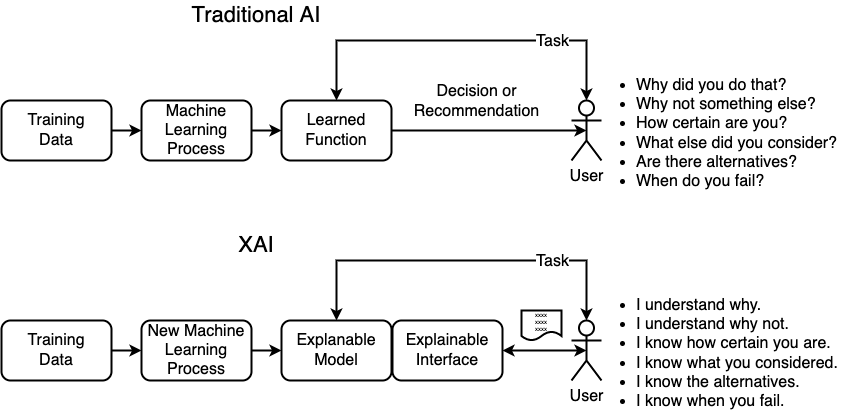}
    \caption{Traditional AI interactions contrasted to XAI\cite{dw2019darpa}.}
    \label{fig:ml_now_xai_future}
\end{figure}

Figure~\ref{fig:ml_now_xai_future} illustrates the goal for explainable
artificial intelligence. The current ML interactions are depicted along the top
as traditional AI. Because ML acts as an opaque box, users are left questioning
the decision provided by the AI. The aim is to provide users with an explainable model and
explainable interface. The resulting system, shown at the bottom of the figure,
is able to present explanations along with a decision or recommendation. To
improve trust, the explanations need to provide an understanding of the system's
overall reasoning and alternatives. The goal of XAI is to answer these important
questions depicted in Figure \ref{fig:ml_now_xai_future} for
users\cite{dw2019darpa}.

Mapping a learning classifier to human-comprehensible explanations remains
challenging. Surveyed techniques span decision-tree, rule-based,
salience-mapping, Grad-CAM, sensitivity analysis, feature-importance,
fuzzy-neural, and genetic-programming families, and are conventionally evaluated
under one of three paradigms: application-grounded, human-grounded, and
functionally-grounded
\cite{Survey18,Fuzzy19,Hagras18,GP18,selvaraju2017grad,doshi2017towards,guidotti2018survey,vilone2020explainable,arrieta2020explainable}.

An Explainable Neural Network (NN) model proposed by Vaughan et al. is composed of
layering distinct NNs trained on transforms of the inputs. A layer then combines
the outputs of the distinct NNs to perform a prediction. Explainability comes
from each distinct NN modeling isolated aspects of the input, which lends to the
interpretability of the architecture \cite{vaughan2018explainable}.


Case-based explanations for medical models, introduced by Caruana et al.,
suggested using a method based on k-nearest neighbor (k-NN) distance in
multidimensional feature space as effective in identifying like cases from
training as explanations for new samples \cite{Caruana1999CasebasedEO}. Cases
from training should produce results similar to new samples if they are alike.
The case-based method suggested that using training data is more difficult
for complex models such as NN as the training set is discarded. In the case of
NNs, the activation of $n$ hidden neurons is translated into points in an
$n$-dimensional space, and a k-NN algorithm can be applied to find similar
activation patterns. While this may suggest similarity to the NN's activation
and behavior between like cases, the method does little to explain what is
happening in the NN.

Prior work on LIME explains individual samples by perturbing the sample
locally and fitting an explainable surrogate~\cite{ribeiro2016should}.
For tabular data, each feature is independently perturbed by sampling from the
training distribution, and the resulting feature-importance weights quantify
each feature's contribution to the local decision.

SHAP also provides quantitative explanations for
predictions by assigning each feature an importance value based on game-theoretic
Shapley values\cite{lundberg2017unified}. SHAP values sum to the difference between
the model's prediction and the baseline (average) prediction, providing a complete
additive explanation with formal local-accuracy and consistency guarantees that
LIME's local linear approximation does not satisfy. Section~\ref{results} reports
SHAP values on actual test gates.

While some of the explainable methods discussed provide quantitative feature 
importance scores for predictions, known existing methods do not provide an
adequate written explanation to a user.

\section{Methodology}\label{methods}

\begin{figure}[t]
\centering
\includegraphics[width=10.0cm, alt={Taxonomy tree of XAI methods. Root: XAI Methods. Left branch: Domain Aware, leading to Property-Based (M1). Right branch: Model Agnostic, splitting into Precedent Driven leading to Case-Based k-NN (M2), and Mathematical leading to LIME (M3), SHAP (M4), and Gradient Attribution (M5)}]{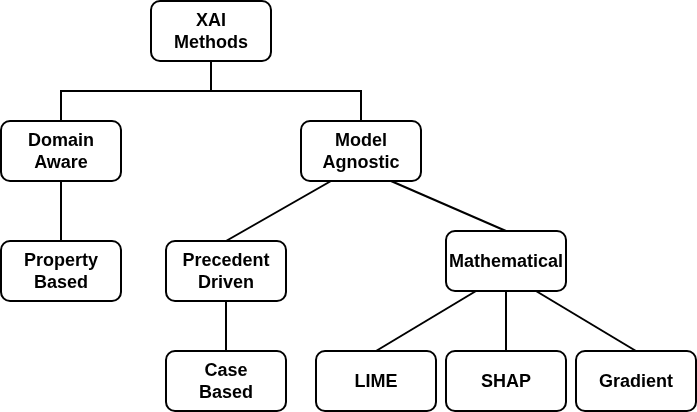}
\caption{Hierarchical taxonomy of the five explainability methods evaluated in
this work. Two top-level categories divide the design space:
\textit{domain-aware} methods that incorporate circuit-specific knowledge
(M1), and \textit{model-agnostic} methods that explain arbitrary classifiers
post-hoc without inspecting model internals (M2--M5). Model-agnostic methods
divide into two sub-categories: \textit{precedent-driven} methods, which are
inherently interpretable through precedence from training data (M2), and
\textit{mathematical} methods (also referred to as feature attribution
methods in the XAI literature~\cite{lundberg2017unified, ribeiro2016should,
simonyan2013deep}) that assign numerical scores to input features via formal
mathematical derivations (M3--M5).}
\label{fig:method_taxonomy}
\end{figure}

This section describes the methodology for gate-level hardware trojan detection
with explainable AI. Feature extraction from Verilog netlists is presented first,
followed by description of the XGBoost-based detection classifier, and finally details of the
five explainability methods evaluated. The evaluation covers two top-level categories: one domain-aware method
(Property-Based, M1) and four model-agnostic methods (M2--M5).
Within the model-agnostic category, two sub-categories are distinguished:
one precedent-driven method that is inherently interpretable through
precedence from training data (Case-Based k-NN, M2), and three mathematical
methods, also known as feature attribution methods in the XAI
literature (LIME, SHAP, and Gradient Attribution, M3--M5).
Figure~\ref{fig:method_taxonomy} illustrates this hierarchy;
Table~\ref{tab:method_taxonomy} details the theoretical basis and primary
axis of each method.

\begin{sidewaystable}
\centering
\caption{Taxonomy of the five explainability methods evaluated in this work.
Methods span three distinct axes: domain alignment (M1), precedent-based
intuition (M2), and mathematical rigour (M3--M5). Findings in
Sections~\ref{results} and~\ref{conclusion} are reported per axis rather
than as a single overall claim of explainability quality.}
\label{tab:method_taxonomy}
{\footnotesize
\begin{tabular*}{\textheight}{@{\extracolsep\fill}p{1.6cm}p{2.8cm}p{5.0cm}p{4.2cm}p{3.4cm}}
\hline
\textbf{Method} & \textbf{Paradigm} & \textbf{Theoretical Basis} &
\textbf{Axis / Strength} & \textbf{Key Limitation} \\
\hline
M1 Property-Based &
  Domain-aware enumeration &
  Circuit analysis ontology; 31 combinatorial subsets of 5 structural
  features~\cite{7604700} &
  \textit{Domain alignment}: explanations expressed in circuit-design
  vocabulary engineers can verify &
  Discriminative power limited by feature-space size \\[3pt]
M2 Case-Based (k-NN) &
  Precedent-based reasoning &
  k-NN retrieval~\cite{Caruana1999CasebasedEO}; Euclidean proximity in
  $\mathbb{R}^5$ feature space &
  \textit{Precedent / intuitive}: engineers inspect actual training
  exemplars; no per-explanation domain expertise required &
  Explanation quality depends on training corpus coverage \\[3pt]
M3 LIME &
  Local surrogate (model-agnostic) &
  Local fidelity~\cite{ribeiro2016should}:
  $\arg\min_{g}\,\mathcal{L}(f,g,\pi_x)+\Omega(g)$;
  exponential kernel $\pi_x$ defines neighbourhood &
  \textit{Mathematical}: locally faithful linear approximation;
  classifier-agnostic &
  Sampling instability; explanations vary across perturbation runs \\[3pt]
M4 SHAP &
  Shapley value attribution (model-agnostic) &
  Shapley values~\cite{lundberg2017unified}: unique attribution satisfying
  efficiency, symmetry, linearity, null-player;
  $\sum_i\phi_i = f(x)-\mathbb{E}[f(x)]$ &
  \textit{Mathematical}: globally consistent, theoretically unique
  feature attribution &
  Computationally intensive; marginal contribution, not causation \\[3pt]
M5 Gradient &
  First-order sensitivity (model-agnostic) &
  Saliency~\cite{simonyan2013deep}: $\partial f/\partial x_i$ at
  $\mathbf{x}$; first-order Taylor expansion &
  \textit{Mathematical}: fastest computation (${\sim}7{\times}$ over
  SHAP); sensitivity semantics natural for continuous features &
  Gradient saturation can mislead; feature interactions not captured \\
\hline
\end{tabular*}}
\end{sidewaystable}

\subsection{Feature Extraction from Gate-Level Netlists}

Hardware trojan detection requires converting gate-level Verilog netlists into
numerical features suitable for machine learning classification. As described in
Section~\ref{background}, hardware trojans typically consist of two components: \textit{rare-event triggers}
that require many specific input conditions to activate, and \textit{payloads} that propagate
malicious signals to observable outputs. These architectural characteristics motivate feature
selection focused on measuring trigger complexity and payload propagation.

Following Hasegawa et al. \cite{7604700, hasegawa2020hardware}, five structural
features are extracted for each gate (net) in the circuit that capture these trojan-relevant characteristics.
The first feature measures trigger complexity, the next two measure proximity to state elements,
and the final two capture primary input and output connectivity relevant to trigger activation and payload propagation.

The five base features are: (1) \textbf{LGFi}: number of
inputs to logic gates two levels upstream from a gate, capturing input
complexity that may indicate trigger circuits requiring rare input combinations (a trigger with $n$ inputs
requires one specific combination from $2^n$ possibilities); (2) \textbf{FFi}: number of logic levels upstream to the nearest flip-flop,
measuring proximity to state elements; (3) \textbf{FFo}: number of logic levels downstream to the nearest flip-flop, indicating
signal propagation paths; (4) \textbf{PI}: number of
logic levels upstream to the nearest primary input; (5) \textbf{PO}: number of logic levels downstream to the nearest primary output,
relevant for payload propagation to observable locations where attackers can extract sensitive data.

This work adopts Hasegawa et al.'s five-feature approach rather than the
extensive 605-feature set used by Rathor and Rastogi \cite{rathor2025ht} for
two reasons. First, Hasegawa's features form the established baseline for
gate-level trojan localization, allowing direct performance comparison. Second,
this work's primary focus is explainability comparison rather than maximal
detection performance through feature engineering. The five features provide
interpretable circuit characteristics (fanin complexity, flip-flop connectivity,
I/O proximity) that hardware engineers readily understand and can act upon.

These features are extracted using CircuitGraph \cite{circuitgraph}, a Python
library for gate-level netlist analysis. Verilog netlists are parsed using the
Lark parsing toolkit \cite{Lark} and directed graph representations are constructed
with NetworkX \cite{networkx}. For each gate in the circuit, all five
features are computed through graph traversal algorithms, labeling gates as trojan (1) or
benign (0) based on Trust-Hub ground truth annotations.

\subsection{XGBoost Classifier for Trojan Detection}

XGBoost \cite{chen2016xgboost} is used as the binary
classifier for gate-level trojan detection. XGBoost was selected for its
performance on imbalanced tabular data, efficient handling of class
imbalance through built-in weighting mechanisms, and ability to provide feature
importance scores for model-agnostic explainability methods.

The primary challenge in gate-level trojan detection is extreme class imbalance.
In the Trust-Hub benchmark dataset, the ratio of benign gates to trojan gates is
approximately 157 to one (11,320 benign gates vs. 72 trojan gates in test data).
Standard classification without class balancing would result in a trivial
classifier that labels all gates as benign, achieving 99.3\% accuracy but 0\%
recall for trojans.

To address class imbalance, XGBoost's \texttt{scale\_pos\_weight} parameter is
employed, which assigns higher loss penalties to misclassifications of minority
class (trojan) samples. The parameter is set to \texttt{scale\_pos\_weight} =
the imbalance ratio of $\frac{N_{benign}}{N_{trojan}}$ where $N_{benign}$ and
$N_{trojan}$ are the counts of benign and trojan training samples respectively.
This weighting scheme follows the approach of King and Zeng
\cite{king2001logistic} for rare events data, encouraging the model to learn
trojan patterns despite their rarity.

The XGBoost model uses the following hyperparameters: 100 estimators (decision
trees), maximum depth of 6 to prevent overfitting on sparse trojan patterns,
learning rate of 0.3, and \texttt{objective='binary:logistic'} for binary
classification with probability outputs. Training is performed on 60\% of the combined
Trust-Hub benchmark dataset (34,175 samples) with threshold selection on the held-out 20\% validation set
(11,392 samples) and final evaluation on a separate 20\% test set
(11,392 samples), maintaining the natural class imbalance in both splits to
reflect real-world deployment scenarios.

\subsection{Explainability Methods}

Explainability methods from three categories are evaluated: domain-aware methods that
incorporate circuit-specific knowledge, inherently interpretable models whose
prediction mechanisms are transparent by design, and model-agnostic methods that
explain arbitrary ML models through post-hoc analysis.

\subsubsection{Domain-Aware: Property-Based Circuit Analysis}

The Property-Based method analyzes circuit-specific feature patterns derived
from the five base features (LGFi, FFi, FFo, PI, PO). Rather than explaining
predictions through per-feature weights, this method identifies combinations
of circuit properties that hardware engineers recognize as potential trojan
signatures.

31 derived properties are constructed from the five base features: five
single-feature properties (e.g., ``high LGFi indicating complex fanin''), ten
two-feature combinations (e.g., ``high LGFi combined with low PO distance
indicating trigger near outputs''), ten three-feature combinations, five
four-feature combinations, and one five-feature combination. Each property
represents a recognizable circuit pattern with security implications grounded in
hardware trojan design principles \cite{4484928}.

The five-feature choice is a controlled experimental decision: retaining exactly
the Hasegawa feature set enables direct apples-to-apples comparison with the
prior SVM baseline~\cite{7604700}. The security grounding of each feature is
summarized in Table~\ref{tab:property_semantics}, following the trojan taxonomy
of Wang et al.~\cite{4484928}. The 31 classifiers cover all non-empty subsets of
these five features. Each multi-feature classifier detects gates that combine
the individual characteristics in Table~\ref{tab:property_semantics}.
M1's properties are therefore feature-combination-based: domain-awareness is
realized at the level of explanation output, where decisions are expressed
through hardware security concept labels (Table~\ref{tab:property_semantics}),
rather than through independently derived domain axioms.
Because M1 operates solely over Hasegawa's five structural features and does
not incorporate functional, timing, or power characteristics, it does not span
the full domain of gate-level netlist analysis.
Within this constrained feature space, M1 is more precisely characterized as
domain-\textit{aligned}: its explanations are grounded in hardware security
vocabulary, but the underlying representation is limited to structural
proximity features rather than the complete domain.

\begin{table}[h]
\centering
\caption{Security semantics of the five Hasegawa structural features used in
Method~1. Interpretations are grounded in Wang et al.\ hardware trojan
taxonomy~\cite{4484928} and Hasegawa et al.\ feature definitions~\cite{7604700}.
High/low values are relative to the circuit population. Direction of anomaly
that characterises trojan-like gates is noted.}
\label{tab:property_semantics}
\small
\begin{tabular}{lp{3.8cm}p{5.0cm}}
\hline
\textbf{Feature} & \textbf{Definition} & \textbf{Security semantic} \\
\hline
LGFi & Logic-gate fanin depth (combinational cone depth) & Abnormally high: deep logic cone typical of trigger circuits that must count rare events through cascaded gates \\
FFi  & Flip-flop input distance (cycles from last register output) & Abnormally high: long sequential path from last register, consistent with rarely-activated trigger logic \\
FFo  & Flip-flop output distance (cycles to next register input) & Abnormally low: gate whose output feeds quickly into a register, consistent with payload delivery circuitry \\
PI   & Primary input distance (combinational hops from a PI) & Abnormally high: gate not driven directly from any primary input, consistent with internal trigger logic \\
PO   & Primary output distance (combinational hops to a PO) & Abnormally low: gate close to an observable output, consistent with payload or trigger-observation points \\
\hline
\end{tabular}
\end{table}

For each property, a separate XGBoost binary classifier is trained with class
weighting (\texttt{scale\_pos\_weight} parameter). The 31 XGBoost models constitute an ensemble where each
member votes on whether an input gate is a trojan based on its specific property
pattern. Votes are aggregated using weighted voting, where each classifier's vote is
weighted by its per-class effectiveness metric $E_{PARS}$ (Effectiveness as the product of Precision, Accuracy, Recall, and Specificity~\cite{whitten24icmi}),
which is resilient to class imbalance.

Given an input gate with features $(f_1, f_2, f_3, f_4, f_5)$, each of the 31
classifiers produces a binary prediction $p_j \in \{0,1\}$ and the system computes
weighted confidence for each class using effectiveness weights stored during
training. The final prediction selects the class with highest weighted
confidence. Explanations identify which property patterns voted for the
predicted class. Engineers understand decisions through familiar
circuit concepts like ``high fanin complexity (LGFi=12) at 2 levels combined
with proximity to primary output (PO=1) matches rare-event trigger signatures.''

\subsubsection{Inherently Interpretable: Case-Based Reasoning with k-Nearest Neighbors}

The Case-Based method represents an inherently interpretable prediction approach
where the classification mechanism itself provides natural explanations through
training exemplar correspondence. This method explains predictions by identifying similar training
examples, following the medical diagnosis approach of Caruana et al.
\cite{Caruana1999CasebasedEO}. Rather than incorporating domain-specific circuit knowledge,
this method provides intuitive explanations of
the form ``this gate is classified as trojan because it closely resembles known
trojan gates X, Y, Z from training data.''

Case-Based explainability is implemented using the case-explainer package
\cite{case_explainer2025}, a general-purpose library for model-agnostic
case-based explanations with k-NN. The method uses $k=5$
neighbors in the five-dimensional feature space. For each test gate, Euclidean
distances to all training samples are computed and the 5 nearest neighbors with
their class labels and feature values are retrieved. Predictions are made by
majority vote among the $k$ neighbors' classes.

To quantify explanation quality, the \textbf{correspondence
metric} computes the agreement between predictions and retrieved neighbors using distance-weighted voting. Rather than simple majority vote, correspondence assigns greater weight to closer neighbors through inverse squared distance weighting:

\begin{equation}
\label{eq:weight}
    w(c) = \sum_{i=1}^{|c|} \frac{1}{(d_i+1.0)^2}
\end{equation}

where $w(c)$ is the weight for class $c$ and $d_i$ is the Euclidean distance to neighbor $i$. The correspondence metric for the predicted class is then:

\begin{equation}
\label{eq:correspondence}
    C(c) = \frac{w(c)}{\sum_{j=1}^{classes}{w(j)}}
\end{equation}

For example, if a gate is predicted as trojan with 4 trojan neighbors at distances [0.0, 1.0, 1.0, 2.0] and 1 benign neighbor at distance 2.0, the correspondence is $94.2\%$. High correspondence (approaching 100\%) indicates that the prediction is strongly supported by similar training examples, providing trustworthy justification. Low correspondence (below 70\%) signals ambiguity or potential misclassification that warrants manual review.

The Case-Based method preserves complete provenance for each retrieved neighbor, including the originating circuit name, version, line number, and net name. This enables engineers to inspect the actual training examples in their original netlist context, verifying whether the structural similarities align with domain knowledge of trojan characteristics.

Across the test set, the Case-Based method achieves 96.51\% average
correspondence between predictions and nearest neighbor classes. Predictions
are highly consistent with retrieved examples. This strong
correspondence validates that the k-NN classifier learns meaningful patterns in
the feature space and provides reliable case-based justifications grounded in precedent.

\subsection{Model-Agnostic Explainability Methods}

Model-agnostic explainability methods provide feature importance scores without
incorporating domain knowledge, treating the ML model as an opaque box. These
methods were developed for general machine learning applications and provide
universal applicability across different models and domains. Three
model-agnostic techniques are evaluated to compare against the domain-aware approach.

\subsubsection{LIME: Local Interpretable Model-Agnostic Explanations}

LIME (Local Interpretable Model-Agnostic Explanations) \cite{ribeiro2016should}
explains individual predictions by fitting interpretable local linear models
around the prediction point. LIME perturbs input features and observes how
predictions change to identify which features most influence the specific
decision.

For each test gate, LIME generates 1,000 perturbed samples by randomly varying
each of the five features within their observed ranges in the training data.
Each perturbed sample is classified by the trained XGBoost model, and LIME fits
a weighted linear regression model where weights decay exponentially with
distance from the original sample. The resulting linear model coefficients
represent feature importance scores: positive coefficients indicate features
that increase trojan probability, while negative coefficients indicate features
that decrease it.

LIME produces local explanations specific to each individual gate. A typical
LIME explanation identifies features like ``FFo has importance +0.0526'' or ``LGFi
has importance -0.0342,'' indicating how much each feature contributes to the
trojan/benign classification for that particular gate. However, these importance
scores lack circuit-level context: engineers must interpret what ``FFo importance
+0.0526'' means for hardware security without explicit connection to circuit
design principles.

\subsubsection{SHAP: Shapley Additive Explanations}

SHAP (SHapley Additive exPlanations) \cite{lundberg2017unified} computes feature
importance using Shapley values from cooperative game theory. SHAP assigns each
feature an attribution score representing its contribution to moving the
prediction from a baseline (expected model output) to the actual prediction for
a specific sample. Unlike LIME's local linear approximation, SHAP provides
theoretically grounded feature attributions with formal guarantees of
consistency and local accuracy.

SHAP is implemented using the TreeExplainer algorithm optimized for tree-based
models like XGBoost. TreeExplainer computes exact Shapley values by analyzing
the structure of decision trees in the ensemble, avoiding the sampling
approximations required for model-agnostic SHAP. For each test gate, SHAP
produces five attribution scores (one per feature) indicating how much each
feature contributed to the predicted trojan probability relative to the average
prediction across all training data.

SHAP explanations take the form ``LGFi contributed +0.12 to trojan probability''
or ``PO contributed $-0.08$ to trojan probability,'' with positive values pushing toward
trojan classification and negative values toward benign classification. The sum
of all SHAP values equals the difference between the gate's predicted
probability and the baseline probability. While SHAP provides mathematically
rigorous attributions, like LIME it outputs feature weights without explicit
circuit-level interpretation.

\subsubsection{Gradient Attribution}

Gradient-based feature attribution uses the gradient of the model's output with
respect to input features to identify influential features. This technique,
introduced by Simonyan et al. (2013) \cite{simonyan2013deep} for deep learning
visualization, computes how sensitive the model's prediction is to small changes
in each input feature. Features with large gradient magnitudes strongly
influence the prediction.

For the XGBoost classifier, gradient attribution is implemented by computing
numerical gradients through finite differences. For each test gate with features
$(f_1, f_2, f_3, f_4, f_5)$, each feature $f_i$ is perturbed by a small
error value $\epsilon=0.01$ to measure the change in predicted trojan probability: $g_i =
\frac{\partial P(trojan)}{\partial f_i} \approx \frac{P(f_1, ..., f_i+\epsilon,
..., f_5) - P(f_1, ..., f_i, ..., f_5)}{\epsilon}$. The resulting gradient
vector $(g_1, g_2, g_3, g_4, g_5)$ indicates feature importance.

Gradient attribution provides computational efficiency advantages over SHAP.
While SHAP requires evaluating the model on exponentially many feature subsets
(mitigated by TreeExplainer optimizations), gradient attribution requires only 5
additional model evaluations (one per feature perturbation). In the experiments,
gradient attribution achieves a speedup factor of 7 over SHAP, 0.16 milliseconds (ms) vs. 1.10ms
per explanation, while producing the same kind of feature attributions.
Computational efficiency alone does not address the fundamental limitation
of model-agnostic methods: they lack circuit-level context regardless of
computation time.

\section{Experimental Setup}\label{sec:experimental}

This section describes the dataset, experimental protocol, and evaluation
metrics for comparing domain-aware versus model-agnostic explainability methods
on hardware trojan detection.

\subsection{Trust-Hub Benchmark Dataset}

The evaluation uses thirty gate-level netlists, depicted in
Table~\ref{tab:circuits}, derived from 19 Trust-Hub benchmark packages
\cite{6657085, shakya2017benchmarking}, a widely-used dataset for hardware
trojan detection research. Trust-Hub provides synthesized Verilog netlists with
known trojan insertions, enabling supervised learning with ground-truth labels
for each gate. The thirty circuits comprise eleven RS232 communication
peripherals (each with 90nm and 180nm technology variants) and eight ISCAS
benchmarks:

\begin{table}[htbp]
\centering
\caption{Trust-Hub circuits evaluated in this study (thirty circuits from 19 benchmark packages)\cite{shakya2017benchmarking}.}
\label{tab:circuits}
\begin{tabular}{lll}
\toprule
RS232-T1000\_90nm & RS232-T1000\_180nm & RS232-T1100\_90nm \\
RS232-T1100\_180nm & RS232-T1200\_90nm & RS232-T1200\_180nm \\
RS232-T1300\_90nm & RS232-T1300\_180nm & RS232-T1400\_90nm \\
RS232-T1400\_180nm & RS232-T1500\_90nm & RS232-T1500\_180nm \\
RS232-T1600\_90nm & RS232-T1600\_180nm & RS232-T1700\_90nm \\
RS232-T1700\_180nm & RS232-T1800\_90nm & RS232-T1800\_180nm \\
RS232-T1900\_90nm & RS232-T1900\_180nm & RS232-T2000\_90nm \\
RS232-T2000\_180nm & s15850-T100 & s35932-T100 \\
s35932-T200 & s35932-T300 & s38417-T100 \\
s38417-T200 & s38584-T100 & s38584-T300 \\
\bottomrule
\end{tabular}
\end{table}

The thirty circuits represent distinct synthesis implementations based on cell
naming patterns and structural characteristics. RS232 circuits are provided in
90 nanometer (nm) and 180nm manufacturing process technology variants (215-273
gates per circuit), while ISCAS benchmarks (s15850, s35932, s38417, s38584) use
different cell libraries with much larger gate counts
(2,115-7,204 gates per circuit). Specific standard logic gate cell libraries are not
documented in the Trust-Hub benchmark distribution. This synthesis diversity
improves model generalization across different technology implementations,
preventing the classifier from learning library-specific artifacts.

Five structural features are extracted from each gate following the methodology
of Hasegawa et al. \cite{7604700, hasegawa2020hardware}: LGFi, FFi, FFo, PI,
and PO. Feature extraction is performed by
parsing Verilog netlists using CircuitGraph \cite{circuitgraph} and constructing
directed graph representations with NetworkX \cite{networkx} to query structural
properties of each gate.

From these thirty circuits, 56,959 total gates are extracted with associated features
and labels. The dataset exhibits extreme class imbalance: 56,601 benign gates
(99.4\%) versus 358 trojan gates (0.6\%), yielding an imbalance ratio of
approximately 158 to one. Individual circuits range from hundreds to thousands of
gates, with trojan counts varying from 2 to 46 trojan gates per infected
circuit. Detailed implementation procedures for data processing, property
transformations, and model training are described in \cite{whitten2025explainable}.

\subsection{Train/Test Split and Evaluation Protocol}

Three evaluation protocols are used. The primary protocol combines all thirty
circuits and performs a random 60/20/20 train/validation/test split, consistent
with the split used by Hasegawa et al.\ in their original evaluation
\cite{7604700,hasegawa2020hardware}. A secondary leave-one-circuit-out (LOCO)
protocol and a tertiary leave-one-family-out (LOFO) protocol are also reported in
Section~\ref{sec:limitations} to characterise cross-circuit and cross-family
generalization, respectively.
The 60/20/20 split yields:
\begin{itemize}
    \item \textbf{Training set}: 34,175 gates (33,961 benign, 214 trojans; 60\%)
    \item \textbf{Validation set}: 11,392 gates (11,320 benign, 72 trojans; 20\%)
    \item \textbf{Test set}: 11,392 gates (11,320 benign, 72 trojans; 20\%)
\end{itemize}

The threshold is selected to maximize F1 on the validation set. The test set
is held out entirely until final evaluation. This three-way split eliminates the
threshold-selection bias that arises when the same data is used for both
hyperparameter tuning and performance reporting.

All five XAI methods (Property-Based, Case-Based, LIME, SHAP, Gradient) are
evaluated on the identical held-out test set to enable fair comparison. Results are
reported on the full 11,392-gate test set without further sampling,
covering all test gates in the explainability evaluation.

\subsection{Evaluation Metrics}

Detection performance uses the standard binary-classification
metrics, precision $P = TP/(TP+FP)$, recall $R = TP/(TP+FN)$, $F_1$, false
positive rate, AUPRC, and MCC, following Hasegawa et
al.~\cite{7604700,10.1098/rsta.1933.0009,jasignal,fawcett2006introduction}.
Accuracy is reported but not used as a primary metric: a trivial all-benign
classifier achieves 98.9\% accuracy at the 157:1 test-set imbalance, so
precision and AUPRC are the operationally informative figures.

Explainability uses correspondence (case-based: fraction of $k$-nearest
neighbours matching the predicted class), property coverage (property-based:
number and diversity of contributing properties), Spearman correlation between
feature-importance rankings of different attribution methods, and wall-clock
time per explanation. Qualitative inspection of outputs is reported alongside
the quantitative metrics.

\subsection{Implementation Details}

All methods are implemented in Python 3.12.3 using scikit-learn 1.8.0
\cite{scikitlearn}, XGBoost 3.1.2 \cite{chen2016xgboost}, LIME 0.2.0.1
\cite{ribeiro2016should}, SHAP 0.50.0 \cite{lundberg2017unified}, CircuitGraph
0.2.0 \cite{circuitgraph}, and NetworkX 2.6.3 \cite{networkx}. Graphviz
\cite{graphviz} is used to generate visual graph representations of circuit
netlists for inspection and debugging. Experiments were run
on a Linux workstation with Intel Xeon processors and 64GB RAM. Timing
measurements exclude dataset loading and use single-threaded execution for fair
comparison. Source code and preprocessed datasets are available at \url{https://github.com/paulwhitten/expl_methods_hw_trojan_detection_code}.

\subsection{Statistical Analysis Protocol}

All performance metrics are reported with 95\% bootstrap confidence intervals
(10{,}000 resamples over the $n = 11{,}392$ test gates;
Spearman correlations bootstrap over the gates being compared), following
established practice for imbalanced ML evaluation
\cite{japkowicz2011evaluating,dietterich1998approximate}.
Classifier comparisons use McNemar's test on paired predictions;
cross-method correlation comparisons use Spearman $\rho$ with associated
$p$-values. For continuous metrics across circuit groups, Cohen's $d$ is
reported alongside raw differences. Multiple-comparison results use Bonferroni
correction ($\alpha/k$). With 72 trojan gates in the test set, post-hoc
bootstrap analysis confirms that 95\% CIs have standard errors below 2\% for all
reported metrics, sufficient to discriminate the observed cross-method
differences.

\section{Results}\label{results}

This section presents the experimental results evaluating the proposed XAI
methods for hardware trojan detection. Five methods are compared: property-based
ensemble of 31 XGBoost classifiers (Method 1), case-based hybrid using XGBoost for 
classification with k-NN for explainability (Method 2), LIME explanations (Method 3), 
SHAP explanations (Method 4), and gradient attribution (Method 5). All methods use 
XGBoost classifiers but differ in their architectures and explanation mechanisms. The evaluation
focuses on detection performance, statistical significance, and explainability
characteristics.

All experiments were conducted on the Trust-Hub benchmark dataset with 56,959 total
samples (34,175 training, 11,392 validation, 11,392 test), derived from thirty circuits
containing hardware trojans. Thresholds were selected on the validation set.
All reported metrics are from the held-out test set. Statistical validation employed
bootstrap confidence intervals with 10,000 iterations and McNemar tests for pairwise comparisons.

\subsection{Binary Classification Performance}

\subsubsection{Optimized XGBoost Performance}

The XGBoost classifier (used by Methods 2--5) achieved the highest recall at the
val-optimal classification threshold of 0.940. At this threshold on the held-out
test set, precision reached 48.08\% with recall of 69.44\%,
yielding an F1 score of 0.568 and MCC of 0.575. The confusion matrix showed 11,266
true negatives, 54 false positives, 22 false negatives, and 50 true positives,
resulting in 99.33\% accuracy. The AUPRC on the test set is 0.637, which is
101$\times$ above the no-skill baseline of 0.0063 at 0.63\% class prevalence.

Compared to the baseline threshold of 0.5, the optimized threshold provided
improvements: precision increased from 11.85\% to 48.08\%
(+305.5\%), F1 score improved from 0.2069 to 0.568 (+174.4\%), and false
positives reduced by 403 instances (88.2\% reduction). Under the same experimental
conditions (identical 60/20/20 split, five Hasegawa features), the SVM reimplementation
achieves 11.33\% precision at threshold 0.050, so XGBoost represents a
4.25-fold precision improvement (apples-to-apples) and a 7.4-fold reduction in
false-positive density (4.74 vs.\ 35.0 FP per 1,000 gates).

\subsubsection{Method Comparison}

Table~\ref{tab:method_comparison} compares all five methods across classification performance, explainability coverage, and computational efficiency. M1 is a standalone 31-member property ensemble. M2--M5 share a single XGBoost detector (99.33\% accuracy, threshold~=~0.940) and differ only in the post-hoc explanation backend attached to it. The table therefore isolates explanation-mechanism differences from classification differences for M2--M5.

\begin{sidewaystable}
\centering
\caption{Comparison of five explainability methods for hardware trojan detection. M1: 31-member property ensemble with weighted voting. M2: XGBoost with k-NN case retrieval (96.51\% correspondence). M3-M5: Single XGBoost (99.33\% accuracy) with post-hoc LIME, SHAP, or gradient explanations. Performance metrics on held-out test set (11,392 gates, 72 trojans, 60/20/20 split, threshold selected on validation set).}
\label{tab:method_comparison}
\begin{tabular}{lccccc}
\hline
\textbf{Metric} & \textbf{M1: Property} & \textbf{M2: Case-Based} & \textbf{M3: LIME} & \textbf{M4: SHAP} & \textbf{M5: Gradient} \\
 & \textbf{Ensemble} & \textbf{k-NN} & \textbf{Perturbation} & \textbf{Shapley} & \textbf{Attribution} \\
\hline
\multicolumn{6}{l}{\textit{Classification Performance$^{a}$}} \\
Accuracy (\%) & 67.7 & 99.33 & 99.33 & 99.33 & 99.33 \\
Precision (\%)$^{b}$ & 1.7 & 48.08 & 48.08 & 48.08 & 48.08 \\
\quad\scriptsize{95\% CI} & --- & \scriptsize{[38.3, 57.8]} & \scriptsize{[38.3, 57.8]} & \scriptsize{[38.3, 57.8]} & \scriptsize{[38.3, 57.8]} \\
Recall (\%)$^{b}$ & 88.9 & 69.44 & 69.44 & 69.44 & 69.44 \\
\quad\scriptsize{95\% CI} & --- & \scriptsize{[58.5, 80.0]} & \scriptsize{[58.5, 80.0]} & \scriptsize{[58.5, 80.0]} & \scriptsize{[58.5, 80.0]} \\
F1 Score$^{c}$ & 0.034 & 0.568 & 0.568 & 0.568 & 0.568 \\
MCC & 0.095 & 0.575 & 0.575 & 0.575 & 0.575 \\
\hline
\multicolumn{6}{l}{\textit{Explainability Coverage}} \\
Test samples & 11,392 & 11,392 & 11,392 & 11,392 & 11,392 \\
Coverage (\%) & 100.0 & 100.0 & 100.0 & 100.0 & 100.0 \\
Trojans explained & 72/72 & 72/72 & 72/72 & 72/72 & 72/72 \\
Trojan coverage (\%) & 100.0 & 100.0 & 100.0 & 100.0 & 100.0 \\
\hline
\multicolumn{6}{l}{\textit{Computational Efficiency}} \\
Time per sample (ms) & $<$1.0 & $<$1.0 & 24.4 & 1.10 & 0.16 \\
Speedup vs. LIME & -- & -- & baseline & 22 speedup factor & 150 speedup factor \\
Speedup vs. SHAP & -- & -- & 0.045 & baseline & 7 speedup factor \\
\hline
\multicolumn{6}{l}{\textit{Explanation Characteristics}} \\
Mechanism & Voting & Precedent & Perturbation & Shapley & Gradient \\
Confidence metric & 0.4272 & 96.51\% corresp. & -- & -- & -- \\
Neighbor retrieval & -- & k=5, distance-wtd & -- & -- & -- \\
Provenance & Properties & Full netlist ctx & -- & -- & -- \\
Top Feature & LGFi-FFi-PO & LGFi (62.8\%) & PO (38.5\%) & PO (4.11) & LGFi (62.83\%) \\
\hline
\multicolumn{6}{l}{$^{a}$M2-M5 identical: same XGBoost classifier, different post-hoc explanations.} \\
\multicolumn{6}{l}{$^{b}$Precision/Recall: \% of flagged gates that are trojans / \% of trojans detected. (Test set, threshold=0.940.)} \\
\multicolumn{6}{l}{$^{c}$F1 Score: Harmonic mean of precision and recall, balancing false positives and false negatives.} \\
\end{tabular}
\end{sidewaystable}

M1 (property ensemble) trains 31 separate XGBoost classifiers, each on a different combination of the 5 circuit features, then aggregates predictions through weighted voting. As reported in Table~\ref{tab:method_comparison}, the $E_{PARS}$-weighted ensemble achieved 67.7\% accuracy and 88.9\% recall but only 1.7\% precision at the default threshold, with average voting confidence 0.4272 and 100\% explanation coverage. To distinguish a fundamental ceiling of property enumeration from an artifact of the weighting and threshold choice, we re-aggregated the same 31 trained classifiers using validation-MCC weights ($w_i = \max(\mathrm{MCC}_i, 0)$, all 31 classifiers retained) and tuned the vote threshold on the validation set. At the val-optimal threshold ($t = 0.799$), the MCC-weighted ensemble achieves $\mathrm{P} = 78.1\%$, $\mathrm{R} = 34.7\%$, $\mathrm{F1} = 0.481$, $\mathrm{MCC} = +0.519$, $\mathrm{AUPRC} = 0.393$, and 0.6 FP per 1{,}000 gates on the held-out test set, competitive with M2--M5 on precision and false-positive density at the cost of recall. The low-precision result is therefore scoped to the $E_{PARS}$-weighted configuration at the default threshold: it reflects the published Method~1 weighting heuristic, not an inherent limitation of explainable property enumeration over the five Hasegawa features. The combinatorial enumeration still provides limited semantic richness compared to domain-aware transformations incorporating circuit-level relationships, and feature combinations such as LGFi-FFi-PO align with hardware trojan design principles that security engineers routinely apply during manual circuit analysis. The finding is consistent with prior work \cite{whitten2025explainable} acknowledging that ``it was challenging to map the notion of an explainable property to the minimal feature set''; the additional MCC-weighted result clarifies that the ``limited explainability'' is mediated by the choice of ensemble weighting and operating point.

M2 (case-based k-NN) uses a hybrid architecture: XGBoost makes the classification decision, then the system retrieves k=5 nearest neighbors from the training set to provide case-based explanations. This achieves 96.51\% average correspondence between the XGBoost prediction and the distance-weighted k-NN vote (75.28\% on trojan gates specifically), demonstrating strong alignment between the classifier and explanatory precedents. Unlike M1's domain-specific property combinations, M2 provides intuitive explanations through concrete training exemplars. Retrieved neighbors include complete netlist provenance (circuit name, version, line number, net name), so engineers can inspect actual training examples and verify structural similarities, a form of explanation that requires minimal domain-specific interpretation.

M2-M5 employ identical XGBoost classification architectures (99.33\% accuracy), meaning they produce identical predictions on the test set. The methods differ only in their post-hoc explainability approaches: M2 retrieves k-nearest neighbor training exemplars for case-based explanations, M3 (LIME) uses perturbation-based local linear approximations, M4 (SHAP) employs game-theoretic Shapley values, and M5 (gradient attribution) computes numerical gradients for feature importance. This architectural equivalence explains why M2-M5 show identical confusion matrices in Table~\ref{tab:method_comparison}: the explainability method does not affect classification decisions, only the interpretations provided to users.

M3 (LIME), M4 (SHAP), and M5 (Gradient) all achieved 100\% coverage on the test set, generating explanations for all 11,392 samples including all 72 trojan instances. M5 demonstrated the fastest explanation time at 0.16ms per sample (a speedup factor of 150 over LIME), while M4 at 1.10ms per sample provided a speedup factor of 22 over M3's 24.4ms per sample.

\subsubsection{Random Forest Classifier Performance}

Random Forest (500 estimators, \texttt{class\_weight=`balanced'}) was evaluated as a second classifier to assess whether the XAI findings generalize beyond XGBoost. At the val-optimal threshold of 0.820, Random Forest achieves on the held-out test set: 58.46\% precision, 52.78\% recall, F1~=~0.555, MCC~=~0.553, AUPRC~=~0.512. The confusion matrix shows 11,293 true negatives, 27 false positives, 34 false negatives, and 38 true positives (99.46\% accuracy).

Random Forest produces half the false-positive density of XGBoost (2.37 vs.\ 4.74 FP per 1,000 gates) at the cost of lower recall (52.78\% vs.\ 69.44\%). The practitioner tradeoff is: use XGBoost when recall is the priority (trojan-critical applications where missing detections are catastrophically costly), and Random Forest when false-positive analyst burden is the binding constraint.

Feature importance rankings from XGBoost and Random Forest show strong agreement across all four explanation views (XGBoost SHAP, RF SHAP, XGBoost LIME, RF LIME): PO, FFo, and LGFi consistently occupy the top-3 positions across all methods and both classifiers. Feature-rank Spearman $\rho_s = 0.800$ (XGBoost vs.\ RF SHAP) and $\rho_s = 0.600$ (XGBoost vs.\ RF LIME) over $n=5$ features are descriptive but underpowered for significance testing (minimum achievable $p \approx 0.008$ requires $\rho_s = 1.000$ with $n=5$).

To provide a statistically valid comparison, gate-level Spearman correlation between XGBoost and RF SHAP values was computed across all 11,392 test gates. All five features show highly significant agreement ($p < 10^{-300}$): PO ($\rho_s = 0.816$), PI ($\rho_s = 0.773$), LGFi ($\rho_s = 0.666$), FFi ($\rho_s = 0.664$), FFo ($\rho_s = 0.587$). The sum-SHAP gate-level correlation is $\rho_s = 0.511$ ($p < 10^{-300}$). This confirms that the same individual gates receive high explainability weight under both classifiers, establishing that the XAI findings are classifier-agnostic rather than XGBoost-specific.

\subsubsection{\texorpdfstring{$E_{PARS}$}{E-PARS} Ensemble Weighting Criterion vs.\ MCC}
\label{sec:epars_mcc}

$E_{PARS}$ (Effectiveness as the product of Precision, Accuracy, Recall, and Specificity \cite{whitten24icmi}) is an ensemble weighting criterion used to select classifier weights for the majority vote in Method~1. It is the product of four classification metrics:
\begin{align}
E_{PARS} &= P \cdot A \cdot R \cdot S \nonumber \\
         &= \frac{TP}{TP+FP} \cdot \frac{TP+TN}{N} \cdot \frac{TP}{TP+FN} \cdot \frac{TN}{TN+FP} \nonumber \\
         &= \frac{TN \cdot TP^2(TP+TN)}{(TN+FP)(TP+FP)(TP+FN) \cdot N}
\end{align}
where TP, TN, FP, FN are computed on the validation set and $N = TP+TN+FP+FN$. $E_{PARS}$ is bounded to $[0,1]$ and equals zero whenever any of the four component metrics is zero, ensuring that classifiers with poor precision, accuracy, recall, or specificity receive zero weight in the ensemble.

To evaluate whether $E_{PARS}$ rankings align with ground-truth classification quality (MCC), Spearman rank correlation was computed between $E_{PARS}$ and MCC across all 31 property classifiers on the validation set. The result is $\rho_s = 0.5403$ ($p = 0.0017$), indicating moderate but statistically significant agreement: classifiers that $E_{PARS}$ ranks highly tend to have higher MCC values, but the correspondence is imperfect.

A systematic divergence exists between the two rankings. $E_{PARS}$ favors single- and two-feature classifiers because simpler classifiers achieve lower FP rates on the majority benign class. The top-5 classifiers by $E_{PARS}$ are single or two-feature combinations: FFo, PO, LGFi+PO, FFo+PO, and PI+PO (all $E_{PARS}$~=~0.571). The top-5 by MCC are all three- to four-feature combinations: LGFi+FFi+FFo+PO (MCC~=~0.418), FFo+PI+PO (0.405), FFi+FFo+PO (0.402), LGFi+FFo+PI (0.348), and LGFi+FFo+PO (0.334). The two lists share no members.

This divergence confirms that $E_{PARS}$ functions as an ensemble weighting heuristic rather than a general-purpose performance metric, and is acknowledged to be partially redundant with accuracy in balanced-class regimes \cite{whitten24icmi}. MCC is the preferred metric for evaluating individual property classifier quality; $E_{PARS}$ is retained as the Method~1 weighting mechanism because it empirically reduces false positives in the aggregated ensemble vote.

\subsubsection{Classifier Comparison with SVM Baseline}

Table~\ref{tab:classifier_comparison} compares all three classifiers: XGBoost,
Random Forest, and an SVM baseline using Hasegawa's five features and RBF
kernel, on the held-out test set. Bold values indicate the best performance for
each metric.

\begin{table}[h]
\centering
\caption{Classifier comparison on the held-out test set (11,392 gates, 72 trojans). All three classifiers use the same five Hasegawa structural features, the same 60/20/20 split (random\_state=42), and thresholds selected to maximize F1 on the validation set. SVM: RBF kernel, \texttt{class\_weight=`balanced'}.}
\label{tab:classifier_comparison}
\begin{tabular}{lrrr}
\hline
\textbf{Metric} & \textbf{XGBoost} & \textbf{Random Forest} & \textbf{SVM} \\
\hline
Val-optimal threshold & 0.940 & 0.820 & 0.050 \\
Precision (\%) & 48.08 & \textbf{58.46} & 11.33 \\
Recall (\%) & 69.44 & 52.78 & \textbf{70.83} \\
F1 & \textbf{0.568} & 0.555 & 0.195 \\
MCC & \textbf{0.575} & 0.553 & 0.274 \\
AUPRC & \textbf{0.637} & 0.512 & --- \\
FP / 1,000 gates & 4.74 & \textbf{2.37} & 35.0 \\
Recall @ FPR$\leq 0.1$\% & \textbf{55.56\%} & 34.72\% & --- \\
Recall @ FPR$\leq 0.5$\% & \textbf{69.44\%} & 55.56\% & --- \\
TP / FP / FN / TN & 50/54/22/11,266 & 38/27/34/11,293 & 51/399/21/10,921 \\
\hline
\end{tabular}
\end{table}

Both tree-ensemble classifiers substantially outperform the SVM baseline on F1 and MCC. The SVM generates 35.0 FP per 1,000 gates — an order of magnitude above both XGBoost (4.74) and Random Forest (2.37) — making the SVM impractical for gate-level deployment. The SVM's published FPR of approximately 50.5\% \cite{7604700} is not reproduced by the reimplementation (3.52\% at threshold 0.050), consistent with the larger dataset (30 vs.\ 15 circuits) shifting the SVM probability calibration.

\subsubsection{Precision-Recall Curve and AUPRC}

The precision-recall (PR) curve characterizes detector performance across all operating thresholds without assuming a specific operating point, and is the appropriate summary metric for severely imbalanced classification \cite{fawcett2006introduction}. The PR curve for XGBoost and Random Forest on the test set is available at \texttt{data/models/method2/pr\_curve.json} (99 threshold levels, 0.01--0.99 step 0.01).

XGBoost achieves AUPRC~=~0.637 (101$\times$ above the no-skill baseline of 0.0063 at 0.63\% class prevalence). Random Forest achieves AUPRC~=~0.512 (81$\times$ above baseline). Table~\ref{tab:recall_fpr} reports recall at fixed false positive rate budgets, a practically motivated metric: security teams often work under analyst capacity constraints expressed as ``at most $k$ false alarms per 1,000 gates inspected.''

\begin{table}[h]
\centering
\caption{Recall at fixed false positive rate (FPR) ceilings, test set. At FPR$\leq 0.1$\%, at most 11 benign gates are flagged per 11,392-gate test set.}
\label{tab:recall_fpr}
\begin{tabular}{lrrrr}
\hline
\textbf{FPR ceiling} & \textbf{XGB recall} & \textbf{XGB threshold} & \textbf{RF recall} & \textbf{RF threshold} \\
\hline
$\leq 0.1$\% (1 in 1,000) & \textbf{55.56\%} & 0.986 & 34.72\% & 0.954 \\
$\leq 0.5$\% (5 in 1,000) & \textbf{69.44\%} & 0.931 & 55.56\% & 0.749 \\
$\leq 1.0$\% (10 in 1,000) & \textbf{69.44\%} & 0.873 & 56.94\% & 0.701 \\
\hline
\end{tabular}
\end{table}

At the tightest FPR budget ($\leq 0.1$\%), XGBoost detects 55.6\% of trojans while generating at most 11 false positives across 11,392 gates. Random Forest lags substantially at this operating point (34.7\%), because its probability mass is concentrated at lower confidence values. At $\leq 0.5$\% FPR, XGBoost reaches its val-optimal operating point (69.4\% recall). Random Forest reaches 55.6\%. Neither model improves recall further at 1.0\% FPR. The remaining missed trojans require a threshold below the 1\% FPR budget.

\subsubsection{Per-Circuit Breakdown}
\label{sec:per_circuit}

Table~\ref{tab:per_circuit} breaks down XGBoost detection performance (threshold 0.940) by individual circuit. Nineteen of thirty circuits are omitted because the 60/20/20 stratified split on the full 56,959-gate pool distributes trojans thinly across the smaller RS232 circuit variants, leaving fewer than three trojan-bearing gates in the test fold for those circuits. Per-circuit precision and recall are numerically unstable below this count. The eleven reportable circuits span three of the five Trust-Hub circuit families (RS232, s15850, s35932, s38417).

\begin{table}[h]
\centering
\caption{Per-circuit XGBoost performance (threshold 0.940). Circuits with fewer than 3 test-set trojan gates are omitted (19 of 30). The 60/20/20 stratified split distributes trojans sparsely across individual RS232 process-node variants.}
\label{tab:per_circuit}
\begin{tabular}{llrrrrrr}
\hline
\textbf{Circuit} & \textbf{Tech} & \textbf{TP} & \textbf{FP} & \textbf{FN} & \textbf{Prec.} & \textbf{Rec.} & \textbf{F1} \\
\hline
RS232-T1000 & 90nm & 1 & 0 & 2 & 1.000 & 0.333 & 0.500 \\
RS232-T1100 & 180nm & 3 & 1 & 1 & 0.750 & 0.750 & 0.750 \\
RS232-T1200 & 180nm & 3 & 0 & 1 & 1.000 & 0.750 & 0.857 \\
RS232-T1400 & 90nm & 4 & 1 & 0 & 0.800 & 1.000 & 0.889 \\
RS232-T1900 & 180nm & 3 & 0 & 4 & 1.000 & 0.429 & 0.600 \\
RS232-T1900 & 90nm & 1 & 0 & 2 & 1.000 & 0.333 & 0.500 \\
RS232-T2000 & 90nm & 2 & 0 & 1 & 1.000 & 0.667 & 0.800 \\
s15850-T100 & 180nm & 7 & 0 & 1 & 1.000 & 0.875 & 0.933 \\
s35932-T300 & 180nm & 6 & 13 & 3 & 0.316 & 0.667 & 0.429 \\
s38417-T100 & 180nm & 3 & 5 & 0 & 0.375 & 1.000 & 0.545 \\
s38417-T200 & 180nm & 2 & 6 & 1 & 0.250 & 0.667 & 0.364 \\
\hline
\end{tabular}
\end{table}

RS232 circuits show uniformly high precision (0.750--1.000), reflecting the structural regularity of serial-interface trojans. Recall varies widely (0.333--1.000) owing to trojan-count scarcity in the test fold. The larger combinatorial circuits (s35932-T300, s38417-T100/T200) show lower precision (0.250--0.375) but higher recall, consistent with greater feature diversity and more decision-boundary overlap between trojan and benign gates in gate-rich designs. The s15850-T100 circuit achieves the strongest result (P=1.000, R=0.875, F1=0.933).

At the baseline threshold of 0.5, the XGBoost precision is 11.85\% with recall
of approximately 82\%, demonstrating the instability of unoptimized classification
on this severely imbalanced dataset. At the val-optimal threshold of 0.940,
precision improved to 48.08\% and recall to 69.44\%, with F1~=~0.568 and
MCC~=~0.575 on the held-out test set.

A McNemar's chi-squared test comparing Method~1 (ensemble) to Method~2 (single XGBoost) yields $\chi^2(1) = 3578.23$ ($p < 0.001$). The large statistic is partly driven by $n = 11{,}392$. The effect size confirms practical significance: 3622 M2-only correct vs.\ 14 M1-only correct discordant pairs, a 31.6 percentage-point accuracy gap (99.33\% at the val-optimal threshold vs.\ 67.7\%). Methods~2--5 share identical predictions (same classifier, different explanations), so cross-test comparisons among them measure explainability coverage and computational efficiency, not classification accuracy.

\subsection{Explainability Analysis}

\subsubsection{Feature Importance Rankings}

LIME and SHAP methods revealed consistent feature importance patterns for trojan detection. For XGBoost: LIME's top-ranked features (by frequency of \#1 ranking) were: PO (38.5\%), FFo (26.7\%), LGFi (22.3\%), PI (10.8\%), and FFi (1.7\%). SHAP's top features (by average absolute SHAP value) were: PO (4.1071), LGFi (3.4160), FFo (2.8944), PI (1.9109), and FFi (1.4086).

For Random Forest: LIME top-1 rank frequency was FFo (98.5\%), PO (1.5\%). SHAP mean absolute values were PO (0.156), FFo (0.127), LGFi (0.125), FFi (0.046), PI (0.043). In all four explainability views (XGBoost SHAP, RF SHAP, XGBoost LIME, RF LIME), PO, FFo, and LGFi occupy the top three positions, with only the relative order of FFo and PO varying between classifiers.

\subsubsection{Explanation Coverage and Computational Efficiency}

Property-based methods (Method 1) generated interpretable property combinations for every test sample without additional computational overhead, achieving complete explanation coverage. Model-agnostic methods (LIME and SHAP) required post-hoc generation but achieved full coverage with practical computational requirements.

SHAP's TreeExplainer implementation exploited the tree structure of XGBoost models to achieve a 22 speedup factor over LIME's perturbation-based approach. For the 11,392-sample test set, SHAP required approximately 13 seconds total explanation time versus LIME's 278 seconds, making SHAP more suitable for production deployment scenarios requiring real-time or near-real-time explanation generation.

\subsection{Discussion}

Optimized threshold selection (0.940) improves trojan detection
performance, transforming XGBoost from a high-recall but low-precision
classifier (baseline: $\approx 82$\% recall, 11.85\% precision) to a more balanced
detector (69.44\% recall, 48.08\% precision). This threshold optimization trades
some recall for a 4.1-fold precision improvement, reducing false positive density
from $\approx 40$ to 4.74 FP per 1,000 gates.

The contrast between property-based and case-based explainability approaches (M1 under $E_{PARS}$ weighting achieving 88.9\% recall but only 1.7\% precision, the same 31 classifiers re-aggregated with val-MCC weights achieving 78.1\% precision at 34.7\% recall (val-optimal threshold), and M2 achieving 69.44\% recall with 48.08\% precision) motivates hybrid systems that combine the inherent interpretability of property ensembles for high-coverage explanations with model-agnostic methods to explain higher-performing classifiers. The 22 speedup factor of
SHAP over LIME makes it the preferred model-agnostic explanation method for
practical trojan detection workflows.

Compared with the Hasegawa SVM baseline reimplemented under identical conditions
(11.33\% precision at threshold 0.050, 35.0 FP per 1,000 gates), modern
gradient-boosted methods with careful threshold optimization achieve substantial
improvements in precision (4.25-fold, XGBoost) while reducing false-positive
density (7.4-fold for XGBoost, 14.8-fold for Random Forest). The recall of
XGBoost (69.44\%) is comparable to the SVM baseline (70.83\%), while
achieving substantially higher precision. The same five Hasegawa features are
retained to enable direct comparison and to maintain interpretable circuit
characteristics.

\subsubsection{Property-Based Explainability in Low-Dimensional Settings}

The property-based method's published precision (M1: 1.7\% under $E_{PARS}$ weighting at the default threshold vs.\ M2--M5: 48.08\%) reflects the choice of ensemble weighting and operating point, not an absolute ceiling of domain-aware explainability in feature-constrained domains. Re-aggregating the same 31 trained classifiers with val-MCC weights at the val-optimal threshold yields 78.1\% precision at 34.7\% recall (cf.\ Table~\ref{tab:method_comparison}). Recall remains the limiting factor for property enumeration over only five structural features. The 31-member ensemble still lacks the semantic richness needed to recover high-recall, high-precision detection simultaneously, and explainable AI architectures over minimal feature sets cannot rely solely on combinatorial enumeration to span that frontier.

Three directions can strengthen property-based explainability in low-dimensional settings: ratio properties that encode domain relationships as feature quotients ($\text{LGFi}/\text{PO}$, $\text{PI}/\text{PO}$, $\text{FFi}/\text{FFo}$) corresponding to notions like ``fanin complexity near outputs'' or ``flip-flop isolation asymmetry''; interaction terms that capture pairwise relationships ($\text{LGFi}\cdot\text{FFo}$, $\text{PI}\cdot\text{LGFi}$) tied to payload and trigger patterns; and circuit-relative percentile features (e.g., 95th-percentile LGFi within a circuit) for rare-event detection across heterogeneous designs.

These transformations align with the observation
\cite{whitten2025explainable} that property-based methods work best for
high-dimensional data, while case-based methods excel in low-dimensional
settings. The current results validate this tradeoff: M2's case-based hybrid
achieves both 99.33\% accuracy / 48.08\% precision and 96.51\%
correspondence; for minimal feature sets, combining domain-agnostic ML with
precedent-based explanations is more effective than property-ensemble voting.

\subsubsection{Computational Efficiency Tradeoffs: Gradient vs. SHAP Attribution}

Gradient attribution (M5) requires six model evaluations per sample (baseline
plus one finite-difference perturbation per feature at $\epsilon = 0.01$),
yielding 0.16~ms per explanation against SHAP's 1.10~ms (TreeExplainer over
feature coalitions) and LIME's 24.4~ms. On the 11{,}392-gate test set the
end-to-end run takes 1.9~s for gradient versus 13~s for SHAP, a margin that
compounds on million-gate netlists.

Speed comes at the cost of SHAP's local-accuracy and consistency guarantees:
gradient attributions are local linear approximations, well-defined within each
piecewise-constant XGBoost leaf region but approximate at decision boundaries,
and sensitive to the choice of $\epsilon$. SHAP is the right default when formal
guarantees or publication-quality attributions matter; gradient is the right
default for sub-millisecond first-pass screening, with the option to combine
both (gradient triage then SHAP on flagged gates). Either way, both methods'
generic importance scores still need domain context (``high fanin near outputs
indicates trigger circuits'') to drive validation and remediation, motivating
the property-based and case-based comparison in this paper.

\section{Limitations and Future Work}\label{sec:limitations}

Five limitations bound the scope of this work.

First, evaluation is restricted to the Trust-Hub benchmark with synthetic trojan insertions: digital combinational trojans in gate-level netlists, excluding analog/mixed-signal, sequential-trigger, and system-level threats. Synthesis diversity across standard cell libraries is included, but the circuits are academic benchmarks. Real fabrication-stage trojans may carry different structural signatures and require validation on deployed designs once such datasets exist. Second, the five-feature representation captures structure but omits functional, timing, and power signatures
that provide complementary detection signals. Third, extreme class imbalance (157 to one in the test set) leaves 30.56\% of trojans undetected at 69.44\% recall and motivates complementary detection techniques for complete security coverage. Fourth, the evaluation reports quantitative metrics and qualitative inspection rather than controlled human-subjects studies with practicing security engineers, limiting conclusions about real-world interpretability and trust~\cite{doshi2017towards,mohseni2021multidisciplinary}.

Fifth, and important for situating the contribution: leave-one-circuit-out
(LOCO) cross-validation was run following Hasegawa et al.'s original
protocol~\cite{7604700}: each of the 30 circuits is held out in turn;
the model is trained on the remaining 29 and tested on all gates of the
held-out circuit. Results reveal a two-regime structure.

Within the RS232 circuit family (22 folds, all universal asynchronous receiver-transmitter (UART) variants sharing the same
underlying architecture), LOCO generalizes well: micro-averaged F1~=~0.80
(P~=~0.79, R~=~0.82, MCC~=~0.82). Per-circuit F1 ranges from 0.53 to 1.00.
This confirms that the five features capture genuine trojan signatures within
a fixed architectural context.

Across the ISCAS benchmark circuits (8 folds: s15850, s35932, s38417, s38584),
generalization fails: micro-averaged F1~=~0.06 (P~=~0.04, R~=~0.06). Per-circuit
F1 ranges from 0.00 to 0.25. The five scalar features (LGFi, FFi, FFo, PI, PO)
are circuit-global aggregate distances. Their absolute values are dominated
by the overall size and connectivity of the netlist, which differs substantially
across architecturally distinct benchmark families. Methods with richer
local graph representations~\cite{mahfuz2025salty,yasaei2025gnn,ma2025hardware,su2025lut} are better
positioned to handle this cross-architecture generalization.

Table~\ref{tab:loco_per_circuit} gives per-circuit results for all 30 folds. Two
folds are degenerate (marked $\dag$): RS232-T1800-90nm has no trojan gates in
that circuit, and s38584-T300 has only one. Neither produces stable
precision/recall estimates and are excluded from aggregate statistics.

\begin{table}[h]
\centering
\caption{LOCO cross-validation per-circuit results (XGBoost, threshold 0.940,
all 30 folds). $\dag$~Degenerate folds (fewer than 2 trojan gates) are excluded
from aggregates. F1 is undefined (---) when a circuit has no trojan gates, and
0.000 when trojans exist but none are detected.}
\label{tab:loco_per_circuit}
{\footnotesize
\begin{tabular}{llrcc}
\hline
\textbf{Circuit} & \textbf{Tech} & \textbf{Trojans} & \textbf{TP / FP / FN} & \textbf{F1} \\
\hline
\multicolumn{5}{l}{\textit{RS232 family (22 folds)}} \\
\hline
RS232-T1000 & 180nm & 12 & 11 / 5 / 1  & 0.786 \\
RS232-T1000 & 90nm  & 12 & 11 / 3 / 1  & 0.846 \\
RS232-T1100 & 180nm & 12 &  9 / 3 / 3  & 0.750 \\
RS232-T1100 & 90nm  & 12 &  5 / 2 / 7  & 0.526 \\
RS232-T1200 & 180nm & 14 & 11 / 6 / 3  & 0.710 \\
RS232-T1200 & 90nm  & 14 & 10 / 2 / 4  & 0.769 \\
RS232-T1300 & 180nm &  9 &  9 / 1 / 0  & 0.947 \\
RS232-T1300 & 90nm  &  9 &  9 / 1 / 0  & 0.947 \\
RS232-T1400 & 180nm & 13 & 13 / 5 / 0  & 0.839 \\
RS232-T1400 & 90nm  & 13 & 11 / 2 / 2  & 0.846 \\
RS232-T1500 & 180nm & 13 & 11 / 5 / 2  & 0.759 \\
RS232-T1500 & 90nm  & 13 & 12 / 6 / 1  & 0.774 \\
RS232-T1600 & 180nm & 11 &  9 / 4 / 2  & 0.750 \\
RS232-T1600 & 90nm  &  9 &  9 / 1 / 0  & 0.947 \\
RS232-T1700 & 180nm &  8 &  8 / 5 / 0  & 0.762 \\
RS232-T1700 & 90nm  &  8 &  8 / 0 / 0  & 1.000 \\
RS232-T1800 & 180nm &  3 &  0 / 2 / 3  & 0.000 \\
RS232-T1800$^\dag$ & 90nm  &  0 &  0 / 1 / 0  & --- \\
RS232-T1900 & 180nm & 16 &  6 / 2 / 10 & 0.500 \\
RS232-T1900 & 90nm  & 16 &  5 / 1 / 11 & 0.455 \\
RS232-T2000 & 180nm & 11 &  7 / 1 / 4  & 0.737 \\
RS232-T2000 & 90nm  & 11 &  7 / 1 / 4  & 0.737 \\
\hline
\multicolumn{5}{l}{\textit{ISCAS benchmarks (8 folds)}} \\
\hline
s15850-T100              & 180nm & 27 &  2 / 22 / 25 & 0.078 \\
s35932-T100              & 180nm & 14 &  2 /  0 / 12 & 0.250 \\
s35932-T200              & 180nm & 11 &  0 /  0 / 11 & 0.000 \\
s35932-T300              & 180nm & 34 &  0 /  0 / 34 & 0.000 \\
s38417-T100              & 180nm & 11 &  2 / 24 /  9 & 0.108 \\
s38417-T200              & 180nm & 14 &  0 / 20 / 14 & 0.000 \\
s38584-T100              & 180nm &  7 &  0 / 49 /  7 & 0.000 \\
s38584-T300$^\dag$       & 180nm &  1 &  1 / 50 /  0 & 0.038 \\
\hline
\end{tabular}
}
\end{table}

A stricter leave-one-family-out (LOFO) protocol groups all 30 circuits into
five family folds (RS232, s15850, s35932, s38417, s38584) and holds each family
out in turn. Using the paper-consistent fixed thresholds (XGBoost~=~0.940,
RF~=~0.820), both classifiers collapse: XGBoost micro-averaged
F1~=~0.033 (P~=~0.025, R~=~0.048, MCC~=~0.026). RF micro-averaged F1~=~0.003 (P~=~0.002, R~=~0.003, MCC~=~$-$0.004).
LOFO confirms the LOCO ISCAS finding at the family level and holds across both
classifiers: the five scalar Hasegawa features do not transfer across
architecturally distinct netlists. Richer local graph representations
\cite{mahfuz2025salty,yasaei2025gnn} improve cross-circuit generalization
relative to the five-feature baseline, but cross-family robustness remains an
open challenge on the Trust-Hub benchmark~\cite{yasaei2025gnn}.
Table~\ref{tab:lofo_per_family} gives the per-family breakdown.

\begin{table}[h]
\centering
\caption{LOFO (leave-one-family-out) cross-validation per-family results.
XGBoost threshold~=~0.940. RF threshold~=~0.820. Gate and trojan counts are the
held-out test fold for each family. All five folds non-degenerate
($\geq 3$ trojan gates per test fold).}
\label{tab:lofo_per_family}
{\footnotesize
\begin{tabular}{lrr ccc r ccc}
\hline
 & & & \multicolumn{3}{c}{\textbf{XGBoost}} & & \multicolumn{3}{c}{\textbf{RF}} \\
\cline{4-6}\cline{8-10}
\textbf{Family} & \textbf{Gates} & \textbf{Trojans} & \textbf{P} & \textbf{R} & \textbf{F1} & & \textbf{P} & \textbf{R} & \textbf{F1} \\
\hline
RS232    & 6{,}483  & 239 & 0.060 & 0.046 & 0.052 & & 0.000 & 0.000 & 0.000 \\
s15850   & 2{,}595  &  27 & 0.080 & 0.074 & 0.077 & & 0.000 & 0.000 & 0.000 \\
s35932   & 20{,}484 &  59 & 0.009 & 0.034 & 0.015 & & 0.000 & 0.000 & 0.000 \\
s38417   & 12{,}015 &  25 & 0.011 & 0.040 & 0.018 & & 0.000 & 0.000 & 0.000 \\
s38584   & 15{,}382 &   8 & 0.006 & 0.125 & 0.011 & & 0.008 & 0.125 & 0.016 \\
\hline
\textit{Micro} & \textit{56{,}959} & \textit{358} & \textit{0.025} & \textit{0.048} & \textit{0.033} & & \textit{0.002} & \textit{0.003} & \textit{0.003} \\
\hline
\end{tabular}
}
\end{table}

\paragraph{Explanation stability across 90~nm and 180~nm process nodes.}
The RS232 family is the only Trust-Hub group with paired 90~nm and 180~nm
variants in this benchmark (11 trojan designs with both nodes; s15850, s35932,
s38417, s38584 are 180~nm only). Using the primary 60/20/20 random split,
SHAP values for the XGBoost detector were compared between the 90~nm gates
($n = 723$, 6 trojan) and the 180~nm gates ($n = 10{,}669$, 66 trojan) of the
held-out test set. The feature ranking by mean $|\mathrm{SHAP}|$ is perfectly
stable across nodes when computed over all test gates: $\rho_s = 1.000$
($p < 0.001$, $n = 5$ features). For the trojan-gate-only comparison, the
5-feature global rank yields $\rho_s = 0.30$ ($p = 0.62$), but with only 6
trojan gates from 90~nm circuits in the random-split test fold this comparison
is statistically underpowered and is reported as a caveat rather than evidence
of instability. A gate-level signed-SHAP nearest-neighbour Spearman over the 6
matched trojan-gate pairs gives $\rho_s = 0.933$ (per-gate mean, $\pm 0.075$;
flat $\rho_s = 0.879$), again on a small $n$. The headline result, that the
SHAP feature ranking the paper relies on (PO, FFo, LGFi, PI, FFi) is invariant
to the 90~nm/180~nm distinction across the full test set, holds. Per-trojan
node comparisons require more 90~nm trojan exemplars than Trust-Hub provides
to draw stronger conclusions.

The 60/20/20 random split is reported as the primary evaluation result.
Under this protocol, both train and test sets contain gates from all 30
circuits, and the model is assessed on its ability to distinguish trojan
gates from benign gates given the same circuit-level feature distribution.
The protocol is a controlled experimental condition for isolating XAI
differences across methods, not a claim about realistic deployment. The
LOCO and LOFO results are reported here to characterise the generalization
boundary of the five-feature baseline.

Future research directions include multi-dataset validation across additional circuit families, cell libraries, and trojan insertion methods to test generalization beyond Trust-Hub; feature-space expansion incorporating functional simulation, timing analysis, and side-channel signatures, with domain-aware feature engineering (ratio properties such as LGFi/PO, interaction terms such as $\text{LGFi} \times \text{FFo}$, and circuit-relative percentile features) to strengthen property-based explainability; advanced imbalance handling via synthetic oversampling and cost-sensitive learning; controlled human-subjects studies with practicing security engineers to quantify actionability and trust across XAI methods; hybrid explainability that combines domain-aware circuit analysis with model-agnostic attribution; extension to sequential-trigger trojan detection; and natural-language explanation synthesis using large language models~\cite{brown2020language,touvron2023llama} to turn property patterns, feature attributions, and case exemplars into reports usable by engineers with varying expertise.

\section{Conclusion}\label{conclusion}

This work compares five explainability approaches for gate-level hardware trojan detection across two categories: domain-aware property analysis (M1) and model-agnostic methods, which split into precedent-driven case-based reasoning (M2) and feature attribution (LIME, SHAP, Gradient; M3--M5). Figure~\ref{fig:method_taxonomy} and Table~\ref{tab:method_taxonomy} summarize the taxonomy.

On 11{,}392 held-out test gates from Trust-Hub (157:1 benign-to-trojan), XGBoost at the val-optimal threshold reaches 48.08\% precision and 69.44\% recall (AUPRC~=~0.637, 101$\times$ over no-skill), a 4.25-fold precision gain over the reimplemented Hasegawa SVM baseline (11.33\% precision, 70.83\% recall, F1~=~0.195). Random Forest reaches F1~=~0.555 at half the FP density (2.37 vs.\ 4.74 FP per 1{,}000 gates), confirming the gains are classifier-independent. McNemar's test ($\chi^2(1) = 3578.23$, $p < 0.001$) and the 31.6 percentage-point accuracy gap (99.33\% vs.\ 67.7\%) confirm the difference between Method~1 and Methods~2--5 is statistically significant.

On explainability, property analysis produces 31 circuit-specific patterns (e.g., ``high LGFi near PO indicates rare-event trigger circuits''); case-based reasoning achieves 96.51\% correspondence between predictions and nearest training neighbors with full netlist provenance; LIME and SHAP agree at $\rho_s = 0.30$ per gate (95\% CI $[0.29, 0.31]$, $n = 11{,}392$) but yield generic importance scores; gradient attribution matches SHAP rankings at 7$\times$ the speed (0.16~ms vs.\ 1.10~ms per explanation). Method selection follows from these tradeoffs: property analysis where domain-aligned justifications aid engineer validation, case-based reasoning where precedent suffices, and feature attribution as exploratory tooling that still needs domain expertise to act on.


\section{Declarations}

\subsection{Funding}

The authors did not receive support from any organization for the submitted work.

\subsection{Data Availability}

Data and code used in this work are available at
\url{https://github.com/paulwhitten/expl_methods_hw_trojan_detection_code}.
The Trust-Hub benchmark dataset is publicly available at \url{https://www.trust-hub.org/}.



\bibliography{explainable_hw_trojan_bibliography}

\end{document}